\documentclass[conference]{IEEEtran}
\IEEEoverridecommandlockouts
\usepackage[numbers]{natbib}
\usepackage{amsmath,amssymb,amsfonts}
\usepackage{algorithmic}
\usepackage{graphicx}
\usepackage{textcomp}
\usepackage{xcolor}
\def\BibTeX{{\rm B\kern-.05em{\sc i\kern-.025em b}\kern-.08em
		T\kern-.1667em\lower.7ex\hbox{E}\kern-.125emX}}

\usepackage{booktabs}

\usepackage{url}

\usepackage{verbatim}
\usepackage{hyperref}

\title{Fiction Sentence Expansion and Enhancement via Focused Objective and Novelty Curve Sampling}

\author{\IEEEauthorblockN{1\textsuperscript{st} Yuri Safovich}
	\IEEEauthorblockA{\textit{Data Science Center} \\
		\textit{Ariel University, Ariel, Israel}\\
	}
	\and
	\IEEEauthorblockN{2\textsuperscript{nd} Amos Azaria}
	\IEEEauthorblockA{\textit{Data Science Center and Computer Science Dept.}} 
	\IEEEauthorblockA{\textit{Ariel University, Ariel, Israel}}
}

\begin{document}
	\maketitle
	\begin{abstract}
		
		We describe the task of sentence expansion and enhancement, in which a sentence provided by a human is expanded in some creative way. The expansion should be understandable, believably grammatical, and highly related to the original sentence. Sentence expansion and enhancement may serve as an authoring tool, or integrate in dynamic media, conversational agents, and advertising.

		We implement a neural sentence expander, which is trained on sentence compressions generated from a corpus of modern fiction. We modify the objective loss function to support the task by focusing on new words, and decode at test time with controlled curve-like novelty sampling.
		We run the sentence expander on sentences provided by human subjects and have humans evaluate these expansions. The generation methods are shown to be comparable to, and as well liked as, subjects' original input sentences, and preferred over baselines.

	\end{abstract}

	\section{Introduction}
	One of the most important skills acquired in elementary school as well as in high-school is the art of writing. While young authors are able to write short and simple sentences, they struggle with writing more complex ones \cite{hilfman1970can}. This is because writing is quite different from speech, and it takes some effort for elementary or even high-school students to develop and write complex sentences. The task of sentence expansion and enhancement, which we describe in this paper, is to take a short sentence provided by a human or an agent, and expand it in some creative way, to a more sophisticated sentence. The new sentence should mostly preserve the content of the original but may not strictly contain the original text (as opposed to infilling). The enhancement may be in clarity, memorability or arbitrary subjective measures, and is distinguished in this task from other sentence expansions such as in textual entailment. Enhancement to a same-length or reduced-length sentence would, if applied, respectively parallel the tasks of paraphrasing and sentence compression, and while we design for expansion, some of our methods are applicable to these related objectives. The expansion is successful if the judge prefers it to the original in its context. That is, the task of sentence enhancement is subjective, and depends very much on human perspective
	
	Sentence expansion and enhancement may serve as an authoring tool, for assisting authors to compose more complex sentences or converting a ``summary'' tell into show, or be integrated in dynamic media, such as games, creating more interesting interactions. Static, human-authored game dialogue systems, for instance, can be likened to curated interaction with conversational agents, and may be an ideal target for a creative writing interface, such as the proposed, as these dialogue systems also do not require advance plot integration planning. Sentence enhancement can also be used by advertising, for creating a variety of options of conveying a specific message, and then picking one or more expanded sentences which are assumed to perform best (e.g., those not shown to this user previously). Finally, conversational agents and virtual assistants \cite{azaria2016instructable,chkroun2019lia,azaria2020agent} and recommender system agents \cite{azaria2016recommender} may use sentence enhancement by generating only the essence of the sentence that needs to be communicated, and this sentence may in turn be expanded to be more human friendly and entertaining.

	For example, given an input sentence such as ``hello world'', a generative model would produce for us ``Oh, hello, a world of peace.'' (This is an actual model output). An ideal model would be better at the task than an average human, and this appears feasible, by our results.

	Computational creativity has always been a popular idea; automatic storytelling and poetry have been attempted from early in computing. Narratology, the study of storytelling, divides it into story (plot) and discourse (style, chronology of presentation) \cite{propp1928morphology}. Most research has been focused on the former \cite{Li2014StorytellingWA}. For example, a set of elements and actions is given, with preconditions and postconditions, and then the construction of a story plot is a search problem. The use of deep neural networks permits generating the language in an integrative way with respect to plot elements. Without some direction, however, the outputs may lack innate meaning. An early example of fully abstractive generation via character-based language model recurrent neural networks (Char-RNN) \cite{Sutskever2011GeneratingTW}:
	``while he was giving attention to the second advantage of school
	building a 2-for-2 stool killed by the Cultures saddled with a half-
	suit defending the Bharatiya Fernall’s office.''
	This artifact is a human-readable metric, a language modeling result used for comparison in neural network research. Although more convincing results appear as the field develops, better grounding for a generative model may be attained by drawing on human participation for input. This enables more nuanced output than in pure generation.
	
	In the literature, most narrative generation methods have been extractive, meaning chosen words or connections are present in some source schema, and logic-, graph- or template-based \cite{Li2013StoryGW,Young1994DPOCLAP,Riedl2010NarrativePB,Chambers2008UnsupervisedLO,Chambers2011TemplateBasedIE}. The topic of sentence enhancement would include slightly modifying words and concepts. In the context of deep learning, abstractive generation is easier, and may be interesting also for the potential relationship with general creativity in AI.	With the recent advances in hardware and neural networks, interest grew in abstractive text generation models, due to capacity for generalization, long-range interactions, and to reduce manual knowledge modeling. A common problem with the essentially statistical systems is generation of safe sequences that are relevant for many inputs \cite{Li2016ADO}. Fully abstractive generation may result in good-looking but meaningless or irrelevant text \cite{Sutskever2011GeneratingTW,Bowman2016GeneratingSF}. Some have mixed neural components in extractive work, and relatively newer models, such as generative adversarial networks (GANs) and variational autoencoders, learn to generate text with diversity through additional parameters in generation or training. However, success relative to standard neural language models is debatable \cite{Semeniuta2018OnAE,Cfka2018EvalAT} for tasks applicable to both.
	At inference, on the decoder side of encoder-decoder sequence to sequence (seq2seq) \cite{Sutskever2014SequenceTS}, beam search tends to produce more generic additions while random sampling is more unreliable, in particular for the task we propose.

	\begin{table}
		\caption{Example sentence expansion and enhancement outputs for human inputs.}\label{tab:ex0}
		\centering
		\begin{tabular}{l|p{6cm}}
			\toprule 
			Input & the woman glared at the child. \\ 
			Output & the old woman glared down at the younger child. \\ 
			\midrule
			Input & the robot looked at jake and smirked, it seemed. \\ 
			Output & the little gray crew looked out at jake and scowled, it seemed like a greatly bad-tempered one.
			\\ 
			\bottomrule 
		\end{tabular} 
	\end{table}
	
	In this paper, models are developed to obtain human input and expand it, transforming complete but possibly unembellished sentences to give them a general or specific style (see Table \ref{tab:ex0}). Given an input (\textit{the robot looked at jake\ldots}), the models add content or context to the input sentence. 
	In terms of the narratological split to story and discourse, the inputs are story clauses. In the example, \textit{robot} expands to \textit{little gray crew} (member?) demonstrating abstractive generation.
	
	We compose a corpus of story sentences paired with their compression (``kernel''). These sentence kernels are obtained by using sentence compression techniques on a corpus of modern (mid to late 20th century) fiction, which is scraped from online resources.
	See Table \ref{tab:exker0} for examples of compression kernels.
	Using this, we train models to transform sentence ``kernels'' to their source form, using the platform of seq2seq with attention \cite{Bahdanau2014NeuralMT} for the expansion task. The seq2seq loss function is first modified to emphasize learning new words, making extended training possible. Then simple alternative test-time sampling methods are applied to better control randomness in decoding. The changes improve output quality, in terms of the average preference of human judges.
	
	\begin{table}
		\caption{Example kernels and original sentences.}\label{tab:exker0}	\centering
		\begin{tabular}{l|p{6cm}}
			\toprule
			Kernel	 & smoke belched from the pipe. \\
			Original & blue smoke belched from the chromed exhaust pipe.
			\\ 
			\midrule
			Kernel	 & are you back? \\
			Original & are you back with the revolutionary lover?
			\\ 
			\bottomrule
		\end{tabular} 
	\end{table}
	
	We run the sentence expander system on crowdsourced human input. It is shown that the best method for sentence expansion results in sentences that are as well-liked by crowdsourced human judges as the input. 
	
	To summarize, the main contributions of this paper are threefold. First, the problem of sentence expansion and enhancement is defined and its importance explained. Second, a method is presented that allows the generation of a large dataset for the sentence expansion problem, by using sentence compression techniques on a given corpus. Capturing the lost information from sentence compression is an approach which to the best of our knowledge is novel. Using compressed sentences does not require any labeled data, and thus allows the construction of data from any source.
	Third, a method is presented for automatic sentence expansion, which is based on several novel ideas, and it is shown that humans prefer sentences produced by the system to original input sentences significantly more often than with baseline systems.

	\section{Related Work}
	
	\label{ss:sentcomp}
	\subsection{Sentence Compression}
	Given an input sentence, sentence compression produces a shorter sentence preserving meaning. Text deletion-based (i.e. extractive) models, used extensively, and newer, abstractive models also, employ word and phrase substitution and reordering, learned from data. A commonly used corpus for abstractive compression or summarization training is Gigaword~\cite{Napoles2012AnnotatedG}, which comprises $\sim$10m documents (4b words) of newswire (with headlines) and auto-generated syntactic annotations. A CNN seq2seq with attention summarizer by \citet{Rush2015ANA}, trained on Gigaword, first-sentence to headline, in their evaluation outperforms an extractive ILP-based (integer linear programming) model by \citet{Clarke2008GlobalIF} and other baselines, although this is likely due to the nature of newswire headlines. \citet{Toutanova2016ADA} compare various metrics, including human evaluators and using four compression systems, and report an opposite relationship on a multi-genre corpus
	wherein ILP is state of the art. ILP as used is a learning algorithm in one form but performance relies on linguistic constraints. \citet{Filippova2013OvercomingTL} report a method for building a parallel corpus for extractive compression from news headlines and first sentences. \citet{Filippova2015SentenceCB} use it to learn LSTM deletion sequences (left to right) with a 2m pair news corpus, reporting 30\% (versus 20\%) perfect match, showing that syntactic features are not required in these DNN models.
	
	\citet{Fevry2018UnsupervisedSC} show controllable-length neural compression without a parallel training corpus by denoising autoencoders, learning to reconstruct a sentence from a list of its words and, as noise, some from another. Desired length is a decoder input.
	
	\subsection{Sentence Generation}
	
	Generation of full text sentences from a mapping, as in translation, may be done with the familiar sequence to sequence (seq2seq) models, which are used in this work as well. Various decoding methods for diversity are found in the field of dialogue generation; however, not many are applicable to the task of expansion. A survey is in \citet{Ippolito2019ComparisonOD}.
	
	One example of non-neural controlled sentence generation is by \citet{ozbal-etal-2013-brainsup}, a framework for slogan generation from keywords, using a dozen feature functions to select words filling slotted patterns. Some of the keywords are special, such as emotion and domain. This is in contrast to arbitrary specification, in language, as in our model.
	
	On the other end, \citet{Fan2018HierarchicalNS} collected a dataset of short ($\sim$700 word) stories written for a sentence premise. They emulate the premise with a GAN generating sentence-length prompts, and then generate stories using convolutional seq2seq, with training-time model fusion \cite{Sriram2018ColdFT}. To support the output length they incorporate gated self-attention heads at different frequencies. This is intended to spread the prompt concepts over many sentences. We train their story expansion seq2seq model on our sentence dataset for a baseline.
	
	Style transfer and conversational models generate sentences from sentences, the latter with context. \citet{Wang2017SteeringOS} use an inverted objective and decode using an MLP sample selector. They focus on a topic by feeding a grid-based topic embedding to the decoder.
	\citet{Prabhumoye2018StyleTT} learn the sentence itself as a latent variable before adversarially generating against style classifiers.
	\citet{Su2018IncorporatingDI} sample sentences in a MCMC process, with a discriminator for constraints such as sentiment. Of research that adapts training for the sampling, \citet{Ranzato2016SequenceLT} optimize sequence decoding based on BLEU or ROUGE scores, using reinforcement learning.
	From a design perspective, \citet{Clark2018CreativeWW} discuss aspects in creative writing with AI, based on slogan generation and story next-sentence suggestion experiments.

	\section{Methods}
	
	\subsection{Focused Objective}
	Using a simple seq2seq model to predict the original sentences from their kernel (the compressed sentence) results in a model that simply copies the input and does not expand it at all.
	This is because all the words that appear in a kernel appear in the original sentence (which is used as the expanded sentence) and merely learning to copy the input provides a relatively low loss. This is clearly not useful.
	Early stopping does not help as it produces an inadequately trained model. However, an incentive could be provided for the model to add additional words not seen in the input, but which appear in the expanded sentence.
	To do this, the negative log-likelihood loss function is modified to increase the importance of learning new words. The cross-entropy of every word in the target (expansion) that is not in the source is multiplied by a factor. After empirically testing values from $2$ to $100$, we chose $10$ to balance effect and convergence. 
	Let $\mathtt{I}$ denote the indicator function for a set, $T$ the ground truth, $S$ the source tokens, and $\lambda=9$ (so that $1+9=10$ for the focusing factor). The modified cross-entropy is thus:
	\begin{equation}
	- \sum_{t} (1+\lambda \mathtt{I}_{T-S}(w_t))\log p(w_t|w_1,\dots,w_{t-1})
	\end{equation}
	
	With this change, the model no longer degenerates to copying its input and can be trained for as long as desired. Preserving words of the input sentence becomes challenging for the model; however, this does not necessarily detract from the goal of improving the input.
	The output becomes longer and more diverse, even without random sampling.
	
	\subsection{Controlled Sampling}
	
	Random sampling is known to give diverse output. However, with softmax temperatures in the most favourable range for our dataset ($0.3-0.7$), in poor expansions we observe arbitrary digressions abruptly halted by the attention search, and on the other end failures of randomness degenerating to the argmax. As expressed in Table \ref{tab:prefs}, different sampling temperatures, beam search, and greedy decoding are empirically equivalent in performance.
	
	Generally speaking, while composing sentences, authors may begin with an intention or idea (which we represent as a sentence kernel) and then decide where to take it or how to write it. Moreover, in a story, the principal plot twists, including the setting setup, tend to occur at significant distance from each other, in order to prevent the audience from becoming fatigued or wary. A conspicuous example of this is the periodicity of cliffhangers in some works. Other examples include intervening segments of comic relief, which may release tension without resolving plot points, and ``filler'' episodes, which pad length. Following this, we examine the concept of a novelty curve, in the simplified story of a single sentence.
	
	We aim for a fixed degree of overall novelty in a sentence, so that it is not affected by sentence length in unstable fashion. In addition, controlling this enables use of models for the above concepts with a fixed count of loci. We calculate per-word novelty as the difference from the softmax maximum. That is, with $p$ and $\tau$ as the probability and corrected sampling temperature at step $t$, 
	
	\begin{equation}
	\begin{aligned}
	nov_w =  \max_y \mathtt{softmax}(\frac{\log p(y)}{\tau}) \\
	- \mathtt{softmax} (\frac{\log p(w)}{\tau})
	\end{aligned}
	\end{equation}
	
	Then, novelty is rationed over expected length, using an accumulator and adjusting the sampling temperature. More complex methods than sampling may be substituted at this point, such as merging some of the top outputs with another layer. To control the rationing several adaptive curves were tested. Each is based on a model of the concept, e.g. novelty that is parabolic, exponential, cyclic, moving window, etc., and adjusts itself (in a step-based decoder) using previous sampling outputs towards a target overall sentence novelty.
	
	The highest-performing model was parabolic. That is, with the middle of the sentence as the center at 0.5,
	
	\begin{equation}\label{eqn:para}
	\tau=b^2 (x-0.5)^2+c
	\end{equation}
	
	This model adjusts the corrected temperature $\tau$ using Eq. \ref{eqn:integral}, where on the left is an integral over $\tau$ for remaining temperature under the curve, with $a$ as the time (current step) divided by expected length, and on the right $t$ is the remaining novelty (target minus accumulated).
	
	\begin{equation}\label{eqn:integral}
	\int_a^1 \left(b^2 (x-0.5)^2+c\right) \, dx=t
	\end{equation}
	
	Solving for one of the parabola's parameters, $b^2$ or $c$, with the other set experimentally as a constant, gives two possible variants for this model, with similar performance.
	
	\begin{subequations}\label{eqn:paras}
		
	\begin{equation}\label{eqn:parab2}
	b^2=\frac{12 (c-a c-t)}{4 a^3-6 a^2+3 a-1}
	\end{equation}

	\begin{equation}\label{eqn:parac}
	c=\frac{-4 a^3 b^2+6 a^2 b^2-3 a b^2+b^2-12 t}{12 (a-1)}
	\end{equation}
	
	\end{subequations}
	
	To assist in aiming for fixed novelty, without retrying until it is within error, some tuning was done as follows. To prevent novelty ``spikes'' at parabola ends unintentionally exhausting the quota too early, top-40 sampling is used, which also reduces irrelevant generation. This can be compensated for by using slightly more powerful parabolas than calculated. For the listed parabolic curves the free hyperparameter is adjusted at design time; other curves can be scaled as needed. UNKs are not generated. To reduce repetitiveness in a simple way, repeated words in a 5-token history are penalized. The settings used are $b^2=0.5$ and $c=3$ for the respective parabolas, repeat penalty is 15 for content words and 10 for stopwords, and the projected expansion factor is 1.65, which is approximately average for both test and corpus expansions. Computationally, these additive steps and the locally contained formulas for temperature resolve in time comparable to random sampling.
	
\begin{figure}[htbp]
		\centering
		\includegraphics[width=\columnwidth]{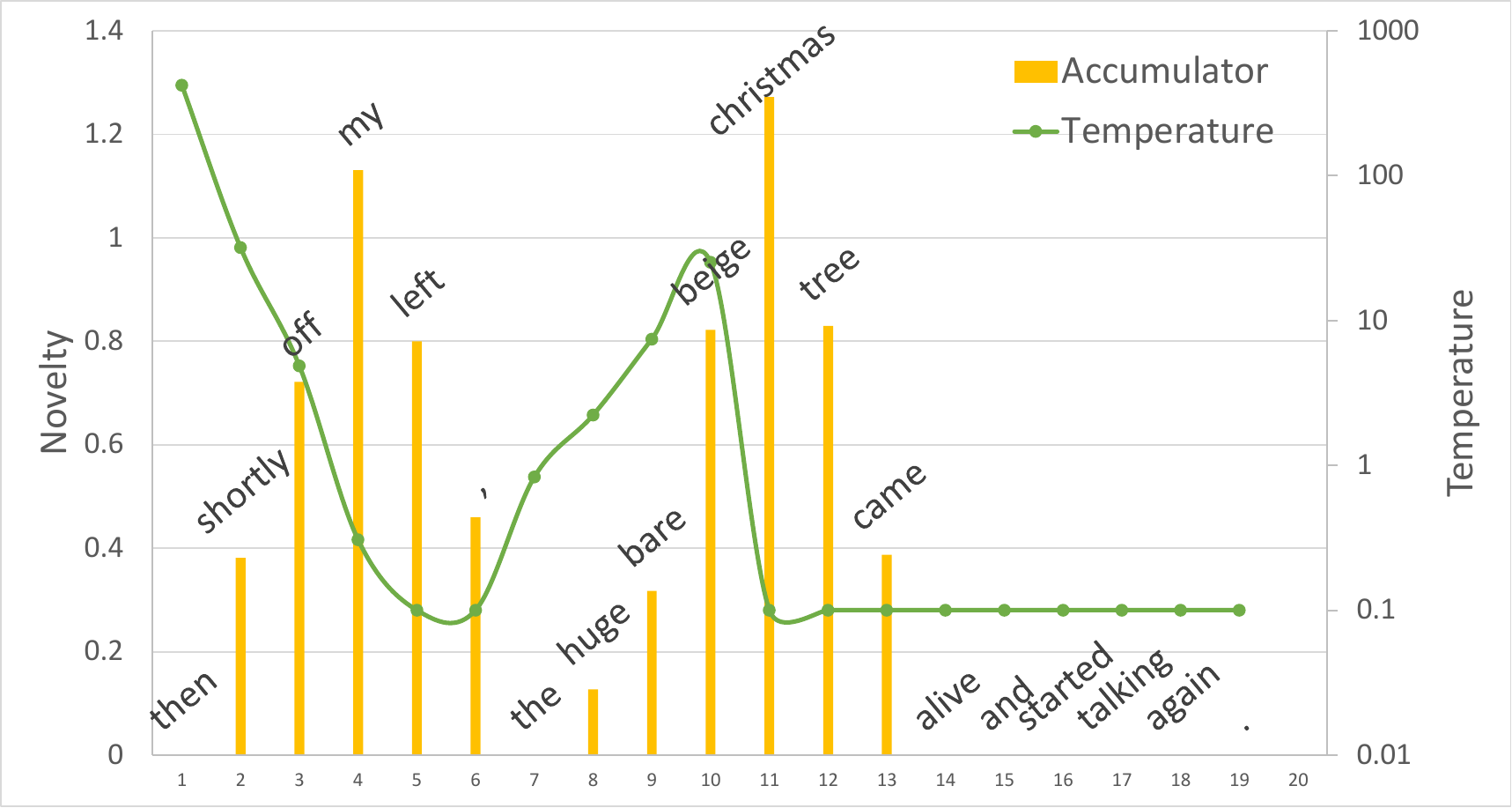}	
		\caption{Decoding the human input ``the tree came alive and started talking" to get ``then shortly off my left , the huge bare beige christmas tree came alive and started talking again". Zoom in for outputs. This instance uses a truncated constant parabola model modified by a size-3 window accumulator (value in orange). Corrected temperature $\tau$ is in green.}
		\label{fig:decoding1}
	\end{figure}

	There may be other parabolic models, or models following other rules, that would be performant. Some less successful models in tests include an exponential on remaining novelty (up to expected length), which spikes $\tau$ somewhere inside the sentence, determined by a coefficient. Another is a windowed accumulator, of 3 or 5 tokens, balancing novelty, following either the target novelty or a parabola. The latter illustrates several concepts in Fig. \ref{fig:decoding1}. As shown, $\tau$, the calculated temperature starts high and words are chosen such that each diverges more from the original sentence: shortly, off, my. After 4 tokens $\tau$ drops as it enters the center of the parabola at half the expected length, producing words that connect the digression with the original sentence: left, comma, the. Near the shallow center \textit{tree} would be a likely token, and would separate the first novelty spike from others. However, in this case $\tau$ quickly rises again producing the words huge, bare, beige, Christmas. At this point the curve is truncated to a limiting value $\epsilon=0.1$. The value in orange shows the accumulated novelty over the past 3 tokens. The accumulator's effect is weak in this case and model, but can be seen on the parabolic curve at steps 4 and 7, where respectively it decreases or increases $\tau$ to compensate for too much or too little novelty.

	\section{Experiments}
	
	\subsection{Corpus} \label{ss:corpus}
	
	Large public corpora of English fiction (e.g. Google Books, Internet Archive) have well-known quality issues with formatting, optical character recognition, and content categorization. One exception is Project Gutenberg, which is proofread. Project Gutenberg's 19th and early 20th-century public domain fiction and nonfiction has dated English, which transfers noticeably in our model. Another corpus used by some authors is BookCorpus, a dump of free user-created fiction from one source. Like several other corpora in use, this has biases, with most of the works being from the romance genre.
	
	For this paper, a corpus was assembled by scraping the Internet for posted content; specifically, identifying and collecting proofread 20th century English fiction by published authors. We believe that this type of cross section better represents the learned Western experience, and can be valuable for sensitive generation tasks. It is hoped that other investigators are likewise inspired to recreate mutable but representative datasets, which are scarce in independent research. Our collection has approximately 600m words in 41m sentences, 45\% of which is speculative fiction.

	\subsection{Optimization}
	In attempts at improving the quality of training, including to reduce the tendency for diverging phrase interpretations and expansions, the corpus was experimentally split into groups by topics. One method used was K-means clustering with a bag-of-words approach. Sentences were word-stemmed 
	and vectorized by either TF-IDF (in this scenario, how specific a word is to its sentence) 
	or hashing. 
	Latent semantic analysis \cite{Deerwester1990IndexingBL} 
	was optionally employed at several dimension parameters (50,100,200,300). The silhouette coefficient of cluster cohesion (defined as the average of scaled point distances to nearest different cluster) was highest (0.621) in the case with 10 clusters (for ``genres''), with LSA to 200 components, 10k features, and counting words showing in 0.001\% to 1\% of sentences. In most cases the clusters were moderately self-similar in appearance and often could potentially be classified as, for example, ``military'', ``bar/pub", ``anatomy'', or ``Star Wars". Such clusters can be used as scenting sets for style priming. Unfortunately, in all cases one cluster was much larger than all others. As the clusters could not be balanced, this approach could not be used to split the corpus. Domain adaptation via other methods, such as in \citet{Axelrod2011DomainAV}, is an avenue for future work.
	
	In a different approach, genre qualifications embedded in the corpus were considered. The romance genre, 7\% of corpus, may be considered fairly homogeneous and a model was trained on this subset, but was not found competitive. Outputs were significantly and perhaps unsurprisingly colored by a focus on relations between objects (whether spatio-temporal or social); this however meant there were fewer novel ``idea'' objects introduced, conflicting with the rationale for abstractive generation, as well as reducing interest for other genre inputs. Since this would limit the possible forms of output, the genre subset was not used further.
	
	\subsection{Setup}
	
	\subsubsection{Compressor}
	
	To generate sentence kernels, the ILP-based system by \citet{Clarke2008GlobalIF} is used. 
	A standard KN-smoothing LM \cite{Kneser1995ImprovedBF} with $1e^-7$ pruning was used for the compression. A supervised model \cite{Rush2015ANA} was also tested using released data. Given the highly restricted nature of news prose which it is trained on, however, in its published configuration this system does not summarize fiction-style sentences convincingly. Inspection revealed a tendency to force a geopolitical framing, and a misunderstanding of common sentence structures in fiction.

	\subsubsection{Data}
	
	The corpus was cleaned of outliers and languages beside English using stopwords and inspection. Text was extracted and preprocessed to segmented sentence form by custom tokenization and segmentation, followed by CoreNLP \cite{manning-EtAl:2014:P14-5}. 
	A large number of exceptional cases in punctuation or style across time, authors, and proofreaders requires that the process is imperfect and some output with excessive-hyphenation was seen, among other issues.
	In the implementation, neutral punctuation, especially quotation marks, often compressed incorrectly. 10\% of sentences are in quotes; 3\% of test set sentences have quotes outside words. Quotes were consequently removed; however, a model trained with them does generate dialogue and narration together.
	Target compression was set to 40\%; average was 31\%. 
	A subset was selected where at least 30\% reduction occurred, in 17 million sentences.
	The use of this set corresponds to the technique of separating short items from a neural model, and has similar observed advantage. This subset is the base for training. 3000 sentences were held for development. Sets were shuffled and lowercased, digits replaced by \#.
	
	\subsubsection{Training}
	
	Models with 4 layers LSTM 1024 encoder and decoder were trained for 1 million steps with 0.2 dropout. Vocabulary size is 50k using SentencePiece \cite{kudo-richardson-2018-sentencepiece}. We note that with many neologisms and domain terms in science fiction, a common genre in this corpus, many ``literary'' words may not appear in a 100k regular vocabulary. Names were not removed, in order that they may be synthesized directly (and thematically). A subword BPE vocabulary \cite{Sennrich2016NeuralMT} tended to produce many nonsense words in experiments; SentencePiece produced less. Sentences over 50 words (1.5\%) had words in excess truncated.
	
	\subsubsection{Test}
	
	A test set of 100 sentences was crowdsourced from 20 workers on Mechanical Turk. Workers were asked to author a ``short sentence that might have appeared in some imaginary story'', with no example given. 
	Statistically, lengths are balanced (with a mean of 12 words, standard deviation 5.2). The mean was affected by the size or width of workers' input text area in multiple rounds of collection. Sentences with profanity or political entities were filtered.
	
	During model evaluation, each input sentence and its expansion (obtained from the model) were compared by 3 unique, high-quality US workers, 
	which were to choose the sentence that they ``think is better or more interesting''. The instructions to the workers were intentionally as short as possible, so that their subjective perspective is captured.
	The workers were not provided any example answers, to minimize researcher bias and let workers weigh brevity against novelty. That is, workers had to select either the input sentence or its expansion, separately for each model.  
	Workers were not asked to rate entailment, but automatic entailment was computed, as described below. Comparisons were shuffled and presented in groups of 5 per worker (for reliability analysis), with random selection order. The workers were \emph{not} instructed that one of the sentences is the initial and one is the expansion (as doing so would clearly bias their responses), even though this would likely be inevitable if paragraph context were given.

	Expansions that failed to terminate within the length limit (50 tokens) or have clearly unnatural repetitiveness, defined as a sequence of length $\geq10$ repeated within a distance of 15, 
	were removed (this happened in about 2\% of the sentences for our method, but more sentences were removed for the baseline methods).
	The number of removals was reduced when penalties for repetitiveness were set.
	
	\subsection{Results}\label{ss:res}

	\begin{table}
		\caption{Sampling method evaluations.}\label{tab:prefs}	\centering
		\begin{tabular}{lcc}
			\toprule
			Sampling	 & Preference		& Significant metrics (if any) \\ 
			\midrule
			Parabola $(c)$  &	 \textbf{0.5}	 \\ 
			Parabola $(b^2)$ & 0.483	 \\ 
			
			Greedy		  &	 0.422		\\
			Random 0.7	  &	 0.417		& Frechet $r=0.26$ \\ 
			Random 0.3	  &	 0.417		 \\  
			Beam search	  & 0.413	\\	
			
			Fan et al. s2s & 0.3	\\
			Trigram freq.  &	 0.1	\\	
			\midrule
			Kernel vs. original		  &	 0.717	\\

		\end{tabular} 
	\end{table}
	
	Table \ref{tab:prefs} presents the main results in terms of human preferences of the expanded sentence over the input sentence. In all models, excepting the kernel vs. original test, the input sentences were the $100$ sentences written by the MTurk workers. It is also noted that only in one case a metric reached statistical significance ($r>0.2$) for a sampling method, Frechet distance (described below) for random sampling.
	The term \textit{Parabola} refers to methods arising from Eq. \ref{eqn:integral}, with $c$ or $b^2$ referring to the variable solved for in Eq. \ref{eqn:paras}. Other custom curves did not perform as well and are not shown.
	
	Baselines include:
	\begin{enumerate}
		\item Using the modified objective: random sampling with specified temperature, beam search (width 10), and greedy search;
		\item 
		Kernels held out from training, with original ``expansions'' (Table \ref{tab:exker});
		\item Inserting a word by sampling LM trigram frequency, up to average rate of expansion by other methods;
		\item The seq2seq fusion model of \citet{Fan2018HierarchicalNS} trained on our dataset for 500k steps, with outputs pruned of repeats in the same way. This is using the default top 10 sampling with temperature 0.8, comparable to other baselines; nonetheless, output length and diversity were relatively significantly random.
	\end{enumerate}
	
	As depicted in the table, the human subjects preferred the original sentences (obtained from the original stories) to the compressed sentences (the kernels) 71.7\% of the time. This is in fact our human-level upper bound, as the expanded sentences were actual story lines.
	Our parabola method outperformed the baselines, and appears to have reached human level equivalence in this scheme, with 50\% of human subjects preferring the expansions to the original human input.
	
	\begin{table*}
		\caption{Example expansions for human input, with preference vs. input (human evaluation).} 
		\label{tab:ex1}	\centering
		\begin{tabular}{c|p{14cm}}
			\toprule
			Input	 &  they were creeping around the corner when they heard a horrible scream.  \\
			Beam search & and then they were rushing around the corner, when they'd first heard a faint scream, and then turned to look at each other's eyes. (33\%) \\ 
			Random 0.7 & and now, in all these other respects, they were both rushing back around the corner, when they'd first heard a strange scream of distress, and then very quietly. (0\%)\\ 
			
			\midrule
			Input	 &  there was a princess that lived in a castle.  \\
			Beam search & but there was also a princess that still lived in a small castle. (33\%) \\ 
			Random 0.7 & but there was also a romance-a queen that lived in a larger, spacious, rambling castle. (66\%)\\ 
			Exponential & now, there was a new, high-ranking and female-american soul that lived in a castle. (66\%) \\
			\midrule
			Input	& she often wondered about it but she did not ask him.  \\
			Parabola $c$ & she'd always wondered about it, but in that case she was not even to ask him.
			(66\%)\\
			Fan et al. s2s & but she'd often wondered about it, but she did not mean him in the first place, but in that kind of way. (33\%) \\
			
			\midrule
			Input	&  the kind found his perfect princess.  \\
			Fan et al. s2s & the kind of family found his perfect princess, the only one. (33\%)\\
			Beam search & the kind of man who 'd found his own, was a good catholic princess. (66\%)\\
			
			\bottomrule
		\end{tabular} 
	\end{table*}

	\begin{table*}
	\caption{Unfiltered example expansions for human input from the Parabola $c$ method.} 
	\label{tab:ex2}	\centering
	\begin{tabular}{l}
		\toprule
smoke was going into the room.   \\ but this white smoke was going down into the empty room anyway. \\  
\midrule
why he was called will, she did not know. \\ why... he -- therefore was called will, and then she did not even know how many times that would be in the first place.\\ 
\midrule
the wizard gave a wry smile at his apprentice.  \\ the dark old lord gave a deep, evil, almost imperceptible female. \\
\midrule
call me jack. \\ please... call me, jack.\\ 
\midrule
this is not her family's crest.  \\ this is certainly not her family's crest, but the other two of my men aren't a good one. \\  
\midrule
she was the luckiest mom in all of the land. \\at this, however she was the most beautiful, ungrateful, and stupid-and yet happy little kid in all of the land. \\ 
\midrule
you are laughing at me, right. \\ but if this is true, then you are laughing at me from this point of view of yours, right. \\ 
\midrule
i want to be a pilot. \\ i do want more than a few minutes, and to be here, a pilot. \\
		\bottomrule
	\end{tabular} 
\end{table*}
	
	Example expansions for human input sentences, with human preference data, are given in Table \ref{tab:ex1}. These examples are chosen to compare across methods, and illustrate user preferences, which are difficult to predict. Unfiltered examples are given in Table \ref{tab:ex2}. Some compression kernels and original sentences of writers (from the corpus) are given in Table \ref{tab:exker}.

	\begin{table*}
		\caption{Example kernels and original sentences, with preference vs. input (human evaluation).}\label{tab:exker}	\centering
		\begin{tabular}{c|p{14cm}}
			\toprule
			Kernel	 &  he put a hand on gabriel's shoulder and guided him. \\
			Original & he put a hand on gabriel's shoulder and guided him from the kitchen and into the shadows of the yard. (100\%)
			\\ 
			\midrule
			Kernel	 & he has feeling for others outside circle of friends and attaches value to life. \\
			Original & he has little feeling for others outside a very small circle of friends , and attaches little real value to human life. (33\%)
			\\ 
			\bottomrule
		\end{tabular} 
	\end{table*}
	
	\subsubsection{Metrics}
	
	Automatic metrics that were tested have low Pearson's $r$ and Spearman's $\rho$ with evaluator preferences ($|r|\leq 0.1$). This varied across sampling methods but generally not to the point of significance.
	
	The metrics computed were:

	(i) Discrete Frechet and cosine distances in InferSent unsupervised sentence embeddings \cite{conneau-EtAl:2017:EMNLP2017};
	
	(ii) ratio of unique added unigrams and bigrams to length (Dist-1 and Dist-2 \cite{Li2016ADO});
	
	(iii) ROUGE-1, ROUGE-2, BLEU-2, BLEU-4 \cite{Lin2004ROUGEAP,Papineni2002BleuAM};
	
	(iv) expansion ratio, added words, and input and output lengths.

	No significant differences in variance or mean were observed in subranges upon plotting of embedding distances and other metrics. Using InferSent, reference MLPs \cite{conneau2018senteval} were trained on SICK dataset entailment and similarity \cite{Marelli2014ASC} and SNLI dataset entailment \cite{Bowman2015ALA}, and again $r$ was negligible. Training the MLPs on preference data for prediction obtained $r=0.02$ on a 10\% test set. Additionally, a relation was manually evaluated on the (100-sentence) beam search results subset, as its relatively generic output may extend to the preservation of meaning (Table \ref{tab:bs17l}). There, for the accurate preservation of entities or concepts (58\% of expansions) $r=0.21$, and for strongly contradicting or changing the meaning (20\%) $r=-0.18$.
	
	\begin{table*}
		\caption{Beam search (width 10) with manually evaluated entailment and InferSent distances.}\label{tab:bs17l}	\centering
		\begin{tabular}{cccccc}
			\toprule
			$r$     & Preserving (58\%) & Contradicting (20\%) & Frechet & Cosine & Dist-1 \\
			\midrule
			Preference &       0.21        &        -0.18         &  -0.1   &  0.18  & -0.12 \\
			Preserving &         -          &          -            &  -0.24  &  0.48  & -0.35\\ 
			\bottomrule
		\end{tabular} 
	\end{table*}

	\section{Discussion}
	
	Many definitions exist for creativity, and this can be source of friction for evaluation methodologies. In one definition, creativity is creating something novel and relevant (for someone or everyone). However, both \textit{novel} and \textit{relevant} may be task-dependent and subjective (for example, are all humans polled? From what age?). Our definition may include utility, while in another field creativity is subsumed by it. Hence, we seek to sidestep the issue of definition by minimizing our own involvement in query interpretation, passing the power to human observers. Then, the most that can be said is that the phrasing "better or more interesting" invokes different perception than a simple "prefer", which might focus more on grammaticality.
	
	It has been argued that a blind evaluation incentivizes generators that are naive of context \cite{Pease2011OnIA}. While we do not ask testers about relevance, in an experiment (not shown), automatic entailment was used to filter result data for those choices where entailment exists, from expansions to original sentences. In this case, the same relative ranking of most generation methods is arrived at, suggesting that their selection does not significantly influence context learning.
	
	Since the task is very subjective in nature, it should not be surprising that raters had some disagreement on $63\%$ of items in evaluations (on average), with chance-adjusted statistics suggesting low reliability in sets.
	The objective for testing was an untouched, unpartitioned measure of quality, and more agreement in low-performing model evaluations would possibly indicate a more uniform set of evaluators. Additionally, distinct NLP tasks exist for other possible elements that may be optimized for, such as sentence compression for brevity.
	
	For sentence expansion, a more relevant or diverse output is not necessarily better. Presumably, each evaluator has different expectations from an expansion, and learning these is important for an authoring tool. The automatic metric results illustrate the necessity of early human evaluation. Preference data given in Table \ref{tab:ex1} shows that preference can be counterintuitive, perhaps due to the diverse population of MTurk workers. In one experiment, two baseline methods, Random 0.3 and trigram frequencies, were compared, and the former was preferred 68.3\% of the time, which may be less than expected given the latter's performance in Table \ref{tab:ex1}.
	
	Expansion outputs sometimes contradict, and the frequency of this is not explained purely by decoding method and the compression removing negators. Given the test collection methodology, input phrases might be inclined towards clich\'e, while in the corpus clich\'e is much likelier to appear in a subverted form. Conversely, conceptually dense sentences such as adages or the already published writing of a veteran author are unlikely to gain from extension (in general style). Splitting off coherence from meaning has not been part of our goal, but grammar and coherence remain significant factors positively correlated with user evaluations in our experiments, whether in input or expansion. This is also reflected in user feedback.
	
	In Table \ref{tab:bs17l} there is correlation of manual entailment with InferSent distances ($r=0.48$ for cosine) on beam search. If an entailment metric is considered reliable, it is easy to resample the output until entailment occurs; this does not seem to affect human preference, however.
	
	Finally, the system allows narrowing the possible style as desired to some degree, using network bias, priming or scenting a pre-trained model with one author's books prior to decoding. A tiny sample is in Table \ref{tab:ex_style}, although generation with specific author styles is not evaluated by humans here.
	
	\begin{table}
		\caption{Example outputs, general and primed style.}\label{tab:ex_style}
		\centering
		\begin{tabular}{l|l}
			\toprule 
			Input & he woke up . \\ 
			Style: \textit{general} & he woke up in the brush. \\ 
			Style: \textit{Douglas Adams} & he woke up, carefully. \\ 
			\bottomrule 
		\end{tabular} 
	\end{table}
	
	\section{Conclusion and Future Work}

	We have defined the task of sentence expansion and enhancement; the task includes a sentence input from humans and requires the production of a more literary, abstractively expanded sentence as output.
	The task is relevant in writing aids, where it may save time and potentially improve quality. Since content is added as well as style, it is relevant for virtual agents and assistants, games, text ads (adding variety), and other media benefiting from adaptable content. 
	A parallel corpus of fiction sentences and their compressions is created, and models are trained on the reverse to perform expansion. A modification to the objective function encourages learning output features and makes training at nontrivial length possible. Simple curve-based sampling methods distribute output novelty in a controlled way. The model outputs, while not independently superior to human inputs, are shown to achieve parity in our evaluation scheme, reaching 50\% of total expansions being preferred, out of an upper bound of 72\% for, essentially, expert human expansions. The result surpasses baselines by 20\% (compared to a sentence adaptation of the model by \citet{Fan2018HierarchicalNS}). 12\% of increase is obtained by the modified seq2seq, and a further 8\% from controlled sampling, in the best performing method. Lastly, it is observed that common metrics of text generation do not predict user preferences for this task.
	
	Our expansions did not compete directly with human generated expansions, and based on experiments the added uncertainty may contribute to a less stable metric, but this second context for comparison across methods, with user post-processing (editing of generated expansions), will be helpful in proving that use of an expander indeed improves quality by saving time. In estimates from our experiments, untrained MTurk writers average one minute to expand a sentence in the same way and to the same length as our system. A complete evaluation suite would include building an assistive user interface, allowing users to choose between different sampling methods and to edit resulting sentences. The interface may learn personal preferences, used as feedback to improve future suggestions.
	Further developments of process may be expected for sentence expansion with paragraph context, and for paragraph expansion and enhancement. Evaluating alternative platforms or improvements to tools in this work is a logical step to surpassing human equivalence in this task setup.

	\section*{acknowledgments}
	This work was supported by the Ministry of Science \& Technology, Israel.

	\bibliographystyle{IEEEtranN}
	
	\bibliography{cit}
	
\end{document}